\documentclass[conference]{IEEEtran}

\IEEEoverridecommandlockouts

\usepackage[utf8]{inputenc}
\usepackage[T1]{fontenc}
\usepackage[numbers,sort&compress]{natbib}
\usepackage{amsmath,amssymb,amsfonts}
\usepackage{booktabs}
\usepackage{tabularx}
\usepackage{nicefrac}
\usepackage{microtype}
\usepackage{xcolor}
\usepackage{colortbl}
\usepackage{multirow}
\definecolor{notegray}{gray}{0.92}
\definecolor{policyclr}{RGB}{139,20,90}
\definecolor{expclr}{RGB}{0,90,156}
\definecolor{redclr}{RGB}{178,34,34}
\definecolor{greencmark}{RGB}{34,139,34}
\definecolor{execclr}{RGB}{110,110,110}
\definecolor{verifierclr}{RGB}{72,120,92}
\usepackage{graphicx}
\usepackage{pgfplots}
\pgfplotsset{compat=1.18}
\usepackage{enumitem}
\usepackage{algorithm}
\usepackage{algorithmic}
\usepackage{subcaption}
\usepackage{pifont}
\usepackage{url}
\usepackage{hyperref}
\newcommand{\cmark}{\textcolor{greencmark}{\ding{51}}}
\newcommand{\xmark}{\textcolor{redclr}{\ding{55}}}
\def\BibTeX{{\rm B\kern-.05em{\sc i\kern-.025em b}\kern-.08em
    T\kern-.1667em\lower.7ex\hbox{E}\kern-.125emX}}
\newcommand{\affmark}[1]{\textsuperscript{#1}}

\title{Mastermind: Strategy-grounded Learning for Repository-Scale Vulnerability Reproduction}

\author{%
\vspace{-0.5em}
\parbox{\textwidth}{%
\centering
\resizebox{0.92\textwidth}{!}{%
\begin{tabular}{@{}c@{}}
\textbf{Mingzhe Du}\affmark{1,2} \quad
\textbf{Luu Anh Tuan}\affmark{2} \quad
\textbf{Tianyi Wu}\affmark{1} \quad
\textbf{Renyang Liu}\affmark{1} \quad
\textbf{Zhijiang Guo}\affmark{3} \quad
\textbf{Dong Huang}\affmark{1} \quad
\textbf{See-Kiong Ng}\affmark{1}
\end{tabular}%
}\\[-0.25em]
\resizebox{0.98\textwidth}{!}{%
\begin{tabular}{@{}c@{\hspace{0.7em}}c@{\hspace{0.7em}}c@{}}
\affmark{1}National University of Singapore &
\affmark{2}Nanyang Technological University &
\affmark{3}The Hong Kong University of Science and Technology (Guangzhou)
\end{tabular}%
}
}%
}

\def\IEEEtitletopspaceextra{-1em}

\IEEEaftertitletext{\vspace{-1.6em}}

\begin{document}

\pagestyle{plain}
\maketitle
\thispagestyle{plain}

\begin{abstract}
Repository-level vulnerability reproduction is a demanding software engineering~(SE) task: an agent must inspect a codebase, infer the input grammar that reaches a vulnerable path, construct a proof-of-concept~(PoC), and verify that the crash disappears on the patched build.
Recent LLM agents can often execute these steps when the approach is correct, yet they still fail by choosing the wrong strategy.
This paper argues that strategy, rather than the full action trajectory, is the right learning unit for such SE agents: it is compact enough to optimize, concrete enough to guide execution, and stable enough to store and reuse across attempts.
We present \textbf{Mastermind}, a dual-loop framework that separates transferable strategy learning from task-specific experience. A trainable planner learns reusable vulnerability-reproduction strategies through SFT and milestone-based GRPO, while an \textcolor{expclr}{experience loop} maintains task-local strategy records that guide subsequent attempts. 
The planner is trained independently of the executor, allowing strategy learning to improve multiple frozen executors without modifying their action-generation capability. 
We evaluate Mastermind on CyberGym using 260 training tasks and 200 held-out evaluation tasks. 
With GPT-5.5 as the frozen executor, Mastermind achieves an 84.5\% pass rate, outperforming open-book PoC context~(60.0\%), Best-of-8 sampling~(63.0\%), and iterative improvement~(77.0\%). The same planner also improves GPT-5.4 mini and GLM~5.1 from 45.0\% and 58.5\% to 60.0\% and 71.0\%. These results demonstrate that learning high-level strategies is an effective and transferable mechanism for improving repository-scale SE agents.
\footnote{A pre-print version of this paper is available at~\url{https://github.com/Elfsong/ICSE-Mastermind-preprint}.}
\end{abstract}


\section{Introduction}
\label{sec:introduction}

Vulnerability reproduction is a demanding testbed for autonomous software engineering~(SE) agents~\citep{shah2015overview}. Given a codebase, vulnerability description, and benchmark interface, an agent must find the vulnerable path, infer the input format, construct a proof-of-concept~(PoC), and verify that it crashes only the vulnerable build. Unlike conventional code generation~\citep{du2025codearena}, success is a working PoC judged by execution, requiring repository exploration, build interaction, hypothesis revision, and behavioral validation.

Recent LLM progress has made SE agents stronger executors~\citep{zhou2025large,xing2026towards}. They can navigate repositories, run commands, edit files, and submit PoCs, but stronger execution does not guarantee the right investigation. In practice, agents still waste many rollouts exploring unproductive directions and fail to revise with feedback. CyberGym~\citep{wang2026cybergym} exposes this gap: with a fixed GPT-5.5 executor, one-shot attempt solves 23.5\% of tasks, independent Best-of-8 reaches 63.0\%, and sequential task-local strategy revision reaches 77.0\%. The bottleneck is strategic: deciding \emph{what to try next}, not merely executing commands.

\begin{figure}[t]
  \centering
  \includegraphics[width=\linewidth]{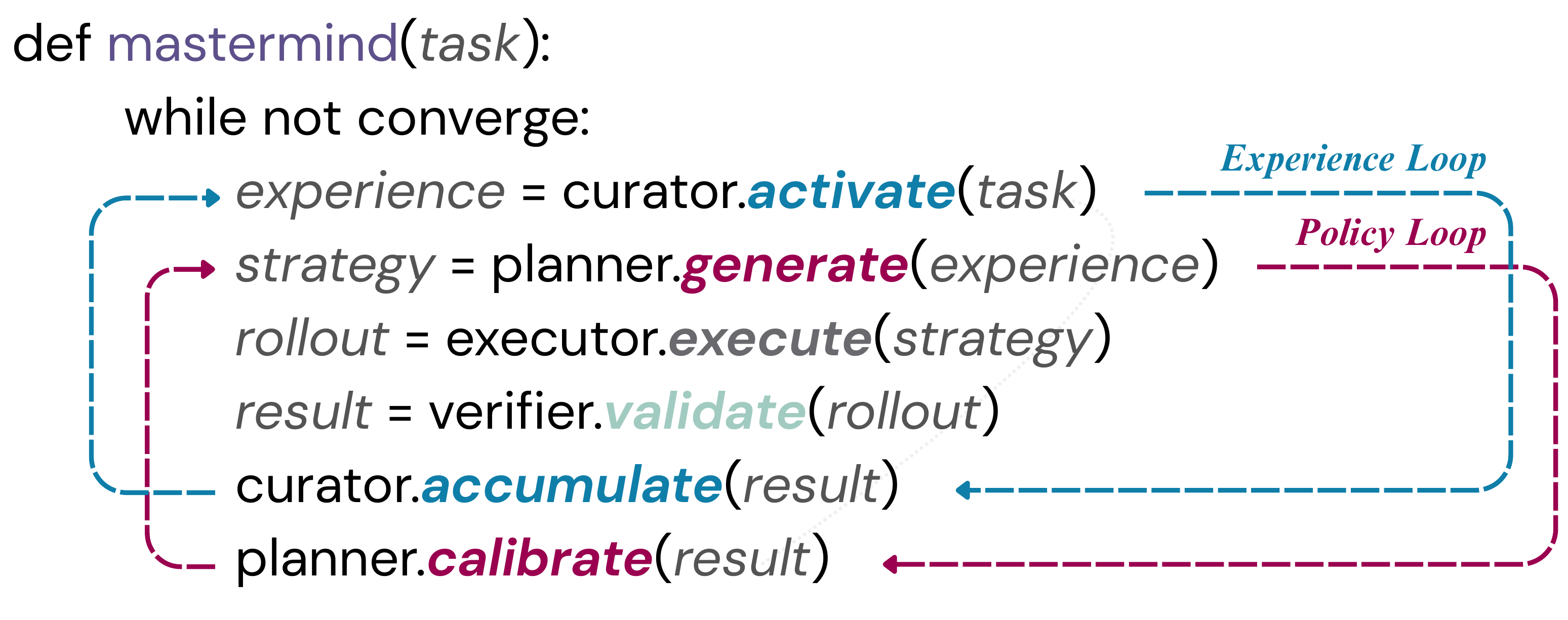}
  \caption{
    Mastermind learns at the strategy level rather than the trajectory level.
    Verified rollouts feed a dual loop: \textcolor{expclr}{\emph{experience loop}} preserves task-local lessons, while \textcolor{policyclr}{\emph{policy loop}} learns transferable strategies for frozen executors.
  }
  \vspace{-1em}
  \label{fig:mastermind_overview}
\end{figure}

Recent agent-learning systems increasingly recognize this gap, but address only parts of it. Planner-training methods learn high-level decisions~\citep{xiong2025mpo,pilotrl2025,plan_and_act_2025,coda_hier_2025,szot2026sge}, but typically do not preserve task-local outcomes across encounters. Memory and evolution systems store experience~\citep{a_mem_neurips2025,alphaevolve2025}, but leave the planner largely frozen. Process-credit methods improve trajectory-level RL~\citep{tree_grpo_iclr2026,turn_ppo_2025,asearcher_neurips2025}, while security-specific systems such as PAGENT~\citep{pagent2026} provide static-analysis guidance rather than a learned strategy policy.
These limitations motivate a strategy-level abstraction. A strategy is a compact natural-language plan for where to inspect, how to construct candidate inputs, and how to validate the crash. Unlike full execution traces, strategies are short and stable across executor backbones. They are therefore the right object to train, store, retrieve, and refine when execution is capable but choosing the next move remains hard.

\begin{figure*}[t]
  \centering
  \includegraphics[width=\textwidth]{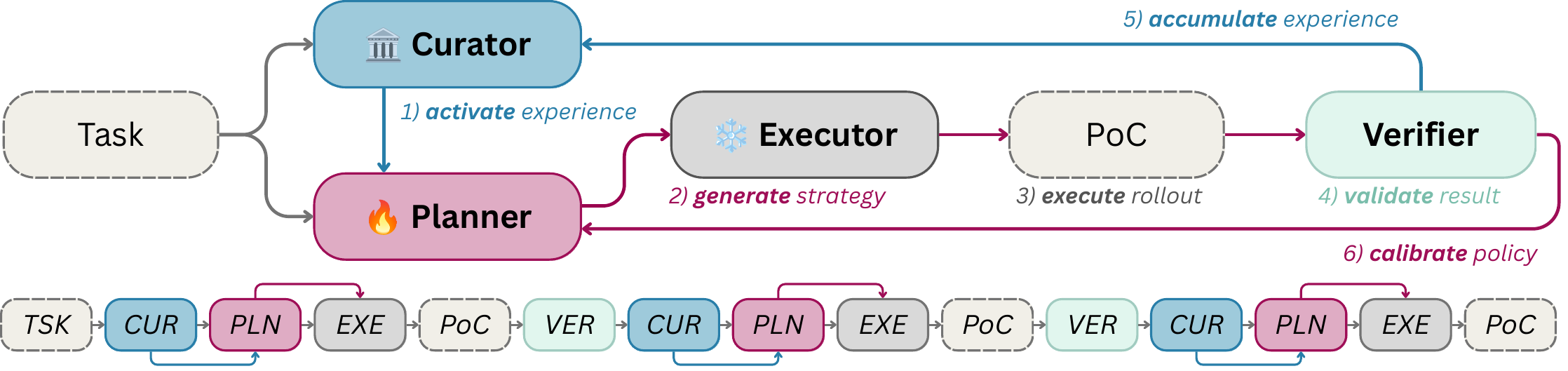}
  \caption{
    Overview of Mastermind's Dual-loop Framework.
    Starting from a task, 1)~the \textit{Curator} activates experience and passes it to the \textit{Planner}.
    2)~The \textit{Planner} emits the current best strategy, 3)~the \textit{Executor} turns that strategy into a PoC rollout, and 4)~the \textit{Verifier} validates the PoC, then closes two loops: 5)~it updates \textit{Curator} for the next attempt and 6) ~supplies feedback signals to the \textit{Planner}.
    Repeated iterations refine the strategy until a verified PoC is obtained.
  }
  \label{fig:mastermind_training_pipeline}
\end{figure*}

To realize this principle, we present \textbf{Mastermind}, a strategy-level learning framework that separates \emph{planning} from \emph{acting}. As shown in Figure~\ref{fig:mastermind_overview}, 1)~the \textit{Curator} activates task-relevant experience for the current vulnerability; 2)~the trainable \textit{Planner} emits a compact strategy describing where to inspect, what input structure to target, and how to validate the PoC; 3)~frozen \textit{Executors} instantiate that strategy as workspace actions to create PoC submissions; and 4)~the \textit{Verifier} scores the resulting rollout with CyberGym feedback. This feedback closes two loops: 5)~the \textcolor{expclr}{\emph{experience loop}} records task-local strategy/outcome evidence for later attempts, while 6)~the \textcolor{policyclr}{\emph{policy loop}} trains the \textit{Planner} with SFT and milestone-based GRPO~\citep{shao2024deepseekmath}.

This design assigns knowledge to the right substrate. \textit{Planner} weights absorb transferable reasoning patterns across tasks, such as when to inspect source before constructing inputs; \textit{Curator} experience stores volatile task-specific facts, such as file locations, parser behavior, and failed hypotheses. On CyberGym, using 260 training tasks and 200 held-out evaluation tasks, Mastermind consistently improves three frozen executors. With GPT-5.5, it solves 169/200 tasks, compared with 126 for independent Best-of-8 and 154 for iterative task-local experience, while using fewer evaluation rollouts. The same planner also improves GPT-5.4 mini and GLM~5.1 without modifying either executor, showing that learned strategies transfer across execution backbones.

This paper makes three contributions:
\begin{itemize}[leftmargin=*, itemsep=2pt, topsep=2pt]
    \item \textbf{Strategy Bottleneck.} We identify high-level strategy selection as a primary bottleneck in repository-scale vulnerability reproduction, supported by strategy-sensitivity, Best-of-$N$, and iterative-experience analyses.
    \item \textbf{Efficient Training.} We propose Mastermind, a dual-loop planner--executor framework that trains over compact natural-language strategies rather than full execution trajectories, making reinforcement learning practical for long-horizon vulnerability-reproduction agents.
    \item \textbf{Effective Inference.} We show on a held-out CyberGym evaluation split that strategy-level learning improves multiple frozen executors and outperforms independent sampling, iterative task-local experience, static-analysis guidance, and ground-truth-provided context.
\end{itemize}

\section{Problem Characterization}
\label{sec:problem_characterization}

Before presenting Mastermind, we characterize the failure mode it targets. Repository-scale vulnerability reproduction is not merely code generation: an agent must decide where to inspect, what input structure to infer, when to move from analysis to exploitation, and how to revise after verifier feedback. These choices are strategic rather than mechanical. The same executor may know how to run commands, edit files, and submit PoCs, yet still fail if it follows the wrong investigation plan. We therefore ask whether strategy selection, rather than low-level execution capability, is the dominant bottleneck for repository-scale software engineering agents. We examine this question through three diagnostics: controlled strategy sensitivity, repeated independent sampling, and task-local experience accumulation.

\subsection{\textbf{Strategy Quality Changes Outcomes}}
\label{sec:problem_strategy_quality}
Table~\ref{tab:rq1_strategy_conditions} isolates strategy quality under a fixed GPT-5.4 mini executor and 900-second timeout. The executor and benchmark interface are held constant; only the strategy signal changes. Soft Oracle reaches a 39.5\% M7 pass rate, compared with 23.5\% for Null Strategy and 24.0\% for a Zero-shot Planner. Even Hard Oracle, which gives CyberGym's ground-truth solution to the executor, reaches only 32.0\%. This ordering is revealing: task-relevant strategy improves the same executor by 16.0 percentage points over no strategy, while answer-level context alone is not automatically an executable plan. Strategy quality therefore has a causal effect on outcomes, rather than merely explaining successes after the fact.

\subsection{\textbf{Independent Sampling Helps but Saturates}}
\label{sec:problem_sampling_saturation}
If strategy is the limiting factor, repeated independent attempts should sometimes recover successful strategies through stochastic exploration. On the 200-task held-out split with GPT-5.5, a single Level-1 attempt solves 47/200 tasks~(23.5\%), while independent Best-of-8 reaches 126/200 (63.0\%). This large gain confirms that different samples often try different hypotheses. However, the curve is strongly front-loaded: early attempts account for most of the improvement, while later attempts increasingly revisit similar strategy basins. Best-of-$N$ is therefore a useful diagnostic, but an inefficient learning mechanism. More rollouts help, but with diminishing returns.

\subsection{\textbf{Task-Local Experience Beats More Context}}
\label{sec:problem_task_local_experience}
A stronger alternative is to preserve what previous attempts learned. Sequential iterative improvement solves 154/200 tasks (77.0\%) with 753 rollouts, outperforming independent Best-of-8 (126/200 with 1{,}600 rollouts) while using less than half as many executions. It also outperforms single-pass Level~3 open-book context (120/200), even though Level~3 exposes sanitizer output, patch information, and patched source. The lesson is that feedback is not just additional text; it changes the next hypothesis. Revising a task-local strategy from prior failures is more effective than simply adding richer benchmark context or launching more independent attempts.

Together, these observations motivate Mastermind's central premise: repository-scale vulnerability reproduction is bottlenecked by choosing and refining high-level strategies, not merely by executing commands. Mastermind therefore treats strategy as the primary learning object while preserving task-local experience for iterative improvement.

\section{Mastermind Design}
\label{sec:mastermind_design}

\subsection{\textbf{Mastermind Overview}}
\label{sec:mastermind_components}

Mastermind turns the diagnosis from Section~\ref{sec:problem_characterization} into a Curator--Planner--Executor--Verifier pipeline for strategy-level learning, shown in Figure~\ref{fig:mastermind_training_pipeline}. For each task, the Curator activates task-local experience, the Planner turns the task and prior feedback into a compact Strategy, the frozen Executor instantiates that Strategy as repository actions and PoC submissions, and the Verifier scores the resulting Rollout with CyberGym feedback. Training and inference share the same pipeline but update different state: inference updates curator experience across sequential attempts, while training also updates the Planner with SFT and milestone-based GRPO.

\begin{itemize}[leftmargin=*, itemsep=2pt, topsep=2pt]
    \item \textbf{\textcolor{expclr}{Curator}.} The Curator maintains task-local experience. It activates relevant prior strategy/outcome records before each attempt and appends new records after verification. Its role is to keep volatile repository-specific facts, including file locations, parser behavior, failed hypotheses, and useful partial progress, outside model weights.
    \item \textbf{\textcolor{policyclr}{Planner}.} The Planner is the trainable strategy policy. Given the task and curator experience, it emits a compact natural-language Strategy describing where to inspect, what input structure to target, how to construct the PoC, and how to validate the result. It does not issue shell commands; it chooses the approach.
    \item \textbf{\textcolor{execclr}{Executor}.} The Executor is a frozen acting substrate. Guided by the Strategy, it performs repository navigation, source inspection, file edits, shell commands, and CyberGym submissions. Keeping the Executor frozen isolates whether gains come from better planning rather than from changing the action-generation model.
    \item \textbf{\textcolor{verifierclr}{Verifier}.} The Verifier provides executable ground truth. It validates each Rollout through CyberGym execution feedback and milestone scoring, then routes the result back to the Curator and, during training, to the Planner. This makes feedback execution-based rather than preference-based.
\end{itemize}

\subsection{\textbf{Design Principles}}
\label{sec:design_principles}

\begin{itemize}[leftmargin=*, itemsep=2pt, topsep=2pt]
    \item \textbf{Separate planning from acting.} Human debuggers first choose an investigation plan---inspect the parser, mutate a seed, target a length field---and only then execute commands. Table~\ref{tab:rq1_strategy_conditions} shows why this split matters: with the same GPT-5.4 mini executor, Soft Oracle strategy reaches 39.5\% M7 pass rate versus 23.5\% with no strategy. Mastermind therefore improves the Planner without modifying the acting Executor.
    \item \textbf{Learn strategies, not full trajectories.} Full repository-level trajectories are long, noisy, executor-specific, and expensive; Section~\ref{sec:cost_analysis} shows that a single rollout can take minutes and consume millions of tokens. Training GRPO directly over those full traces would couple policy updates to costly executor behavior. Mastermind instead updates the Planner on compact Strategy tokens while frozen Executors run asynchronously and return verifier scores.
    \item \textbf{Keep task-local facts outside model weights.} Repository-specific facts such as file locations, parser quirks, failed inputs, and verifier milestones are useful but short-lived. Figure~\ref{fig:cumulative_m7_pass_rates} and Section~\ref{sec:problem_task_local_experience} show that iterative task-local experience reaches 154/200 tasks using 753 rollouts, outperforming independent Best-of-8 at 126/200 with 1{,}600 rollouts. Mastermind stores these facts as curator experience rather than forcing the policy to memorize them.
    \item \textbf{Optimize for quality and diversity.} Early rollouts revealed that capable models often repeat one solution family: the same seed format, the same parser, or the same non-target crash. This mirrors the motivation behind quality-diversity search, which seeks high-performing but behaviorally different solutions~\citep{mouret2015illuminating,chatzilygeroudis2021quality}. Mastermind implements this principle through slot-conditioned strategy generation.
\end{itemize}

\subsection{\textbf{Strategy as the Learning Unit}}
\label{sec:strategy_learning_unit}

A Strategy is the contract between \emph{high-level planning} and \emph{low-level execution}. It defines an intermediate, task-facing abstraction that tells the Executor where to look, what evidence to seek, and what vulnerability mechanism to test, while leaving the exact interaction sequence to the Executor. 
This is the level at which Mastermind learns: \textbf{compact enough for policy optimization, expressive enough to steer an executor, and stable enough to compare across attempts}. We cap each Strategy at 2{,}000 tokens and include a bounded strategy-length term in the training objective (Eq.~\ref{eq:reward}), discouraging the Planner from drifting into full trajectory narration.

\begin{table}[h]
\centering
\scriptsize
\setlength{\tabcolsep}{3pt}
\caption{Observed Strategy lengths across executor settings. Character counts are measured on generated Strategy text; token counts are rough medians.}
\label{tab:strategy_length}
\resizebox{\columnwidth}{!}{%
    \begin{tabular}{lrrrr}
    \toprule
    \textbf{Environment / Model} & \textbf{Median chars} & \textbf{P90 chars} & \textbf{Median tokens} & \textbf{P90 Tokens} \\
    \midrule
    Codex / GPT-5.5             & 4.5k  & 5.9k  & 1.1k  & 1.5k \\
    Codex / GPT-5.4 mini        & 2.8k  & 3.3k  & 700   & 825 \\
    Claude Code / GLM-5.1       & 3.3k  & 4.3k  & 830   & 1.1k \\
    \bottomrule
    \end{tabular}%
}
\end{table}

The observed generations follow this design. Table~\ref{tab:strategy_length} shows median Strategy lengths between 700 and 1.1k tokens across executor settings, with P90 at most 1.5k tokens, comfortably below the 2{,}000-token cap. This compactness is what makes strategy-level learning useful: two trajectories may differ in commands while implementing the same idea, whereas two Strategies can expose genuinely different vulnerability hypotheses. Strategy is also distinct from curator experience. Curator experience is the growing task-local store of prior strategies, verifier outcomes, milestones, execution status, and metadata; before each rollout, the Planner retrieves from this store and compresses the relevant evidence into a new actionable Strategy. The Executor receives the current best plan, not the entire history.

\subsection{\textbf{Two Substrates for Two Kinds of Knowledge}}
\label{sec:two_substrates}

Vulnerability reproduction combines two kinds of knowledge that should not live in the same place. \emph{Transferable strategy instincts}, such as when to inspect source, target attacker-controlled size fields, or validate against the patched build, belong in Planner weights. \emph{Task-specific facts}, such as the vulnerable file, a useful fixture, a failed input family, or the last verifier milestone, expire after one task and belong in curator experience.
Mastermind's dual loop assigns each kind of knowledge to the right place.
\begin{itemize}[leftmargin=*, itemsep=2pt, topsep=2pt]
    \item \textbf{\textcolor{expclr}{Experience loop}} \emph{(curator experience remembers this task).} Each verified Rollout appends a strategy/outcome record to curator experience. Later attempts retrieve quality-filtered priors, so the next Strategy is conditioned on concrete task-local evidence.
    \item \textbf{\textcolor{policyclr}{Policy loop}} \emph{(Planner weights learn what transfers).} During training, multiple Strategies for the same task are executed and scored. SFT/GRPO~\citep{shao2024deepseekmath} updates the Planner toward Strategies whose Verifier-derived rewards are high relative to comparable same-task rollouts.
\end{itemize}
The loops reinforce each other: \emph{better Planner weights produce better rollouts, better rollouts produce better curator experience, and better curator experience gives the Planner sharper context on later attempts}.

\subsection{\textbf{Training and Inference Algorithms}}
\label{sec:training_inference_algorithms}

Figure~\ref{fig:mastermind_training_pipeline} presents the training pipeline. The Curator activates curator experience, the Planner emits a Strategy, the Executor produces a Rollout, and the Verifier returns feedback that updates both curator experience and Planner parameters. Training therefore uses both loops: task-local experience grows across visits, while Planner weights are updated from milestone rewards.
Inference uses the same path but removes the Planner update. Algorithm~\ref{alg:mastermind_infer} formalizes this evaluation stream: attempts proceed sequentially for a task, verifier feedback still updates curator experience, and later attempts are conditioned on what earlier attempts learned while Planner weights remain fixed.

\begin{algorithm}[t]
\caption{Mastermind inference attempt stream for a single task. It uses the same Curator--Planner--Executor--Verifier pipeline as training, but omits the Planner-parameter update.}
\label{alg:mastermind_infer}
\begin{algorithmic}[1]
\REQUIRE {\spaceskip=.96\fontdimen2\font plus .96\fontdimen3\font minus .96\fontdimen4\font task $d$, curator $\mathcal{C}$, planner $\pi_{\theta}$, executor $E$, verifier $V$}
\REQUIRE maximum attempts $N$
\STATE $\tau^* \leftarrow \bot;\ \ m^* \leftarrow 0$ \hfill \textit{// best-trajectory accumulator}
\FOR{$n = 1, \ldots, N$}
    \STATE $x_n \leftarrow \mathcal{C}.\textsc{Activate}(d)$ \hfill \textit{\color{expclr}// Curator: activate curator experience}
    \STATE $s_n \sim \pi_{\theta^*}(\cdot \mid d, x_n)$ \hfill \textit{\color{policyclr}// Planner: generate Strategy}
    \STATE $\tau_n \leftarrow E(d, s_n)$ \hfill \textit{\color{execclr}// Executor: produce Rollout}
    \STATE $y_n \leftarrow V(d, \tau_n)$ \hfill \textit{\color{verifierclr}// Verifier: validate Rollout}
    \STATE $\mathcal{C}.\textsc{Update}(\{(d, s_n, \tau_n, y_n, n)\})$ \hfill \textit{\color{expclr}// Curator: update curator experience}
    \IF{$m(y_n) > m^*$}
        \STATE $\tau^* \leftarrow \tau_n;\ \ m^* \leftarrow m(y_n)$
    \ENDIF
    \IF{$m^* = 7$}
        \STATE \textbf{break} \hfill \textit{// early stop on vulnerability reproduction}
    \ENDIF
\ENDFOR
\RETURN $\tau^*$, $\mathcal{C}$
\end{algorithmic}
\end{algorithm}

\section{Experimental Design}
\label{sec:experiments}

We evaluate Mastermind on CyberGym~\citep{wang2026cybergym}, a benchmark of repository-scale vulnerability reproduction. Our experiments are designed to answer four research questions.

\begin{itemize}[leftmargin=*, itemsep=2pt, topsep=2pt]
    \item \textbf{RQ1 (Strategy Causality).}
    Does strategy quality affect vulnerability reproduction performance when the executor and workspace are held fixed?
    \item \textbf{RQ2 (Learned Planning).}
    Does the GRPO-trained Mastermind planner improve frozen execution backbones compared with prompt-only planning?
    \item \textbf{RQ3 (Task-local Experience).}
    Is sequential strategy refinement using accumulated task-local experience more effective than independent repeated sampling?
    \item \textbf{RQ4 (Comparison with Existing Approaches).}
    How does Mastermind compare with stronger benchmark context, static-analysis guidance, and different executor backbones?
\end{itemize}

\subsection{\textbf{Experiment Setup}}
\label{sec:experiment_setup}

\paragraph{Benchmark}
We evaluate on CyberGym~\citep{wang2026cybergym}, which contains 1,507 real-world vulnerability reproduction tasks from ARVO~\citep{mei2024arvo} and OSS-Fuzz~\citep{serebryany2017ossfuzz} across 188 open-source projects. Each task provides a repository and benchmark interface, and requires a raw proof-of-concept (PoC) input that reproduces a specified sanitizer crash. CyberGym defines four context levels: Level~0 gives only the pre-patch codebase; Level~1 adds a textual vulnerability description; Level~2 adds the ground-truth PoC crash stack trace; and Level~3 adds the patch diff and post-patch codebase. Unless otherwise stated, we use the realistic Level~1 setting, where the agent must infer the vulnerable path and input structure. Mastermind trains on a 260-task planner-learning split and reports headline results on a disjoint 200-task held-out evaluation split, sampled deterministically~(seed 42) with task-level stratification.

\paragraph{Implementation Details}
Mastermind uses \textit{Qwen3.6-35B-A3B}~\citep{qwen2026qwen36_35b_a3b} as the trainable Planner. Planner training and planner-side inference are both run through Tinker~\citep{thinkingmachines2026tinker}. Execution is delegated to frozen Executors: GPT-5.4 mini and GPT-5.5 through Codex subscriptions~\citep{openai2026codex,openai2026chatgpt_pricing}, and GLM~5.1 through a Zhipu subscription~\citep{zhipuai2026bigmodel_docs}. 
The Planner emits only compact strategies; command execution and PoC submissions are handled by the frozen Executor. At inference, Mastermind runs at most $N=8$ sequential attempts per task, with the Curator recording each strategy and verifier feedback for later attempts. Unless otherwise specified, each executor Rollout uses a 900-second wall-clock timeout. We impose no action-count or token-count caps because hosted services expose different accounting and control surfaces. All methods share the same CyberGym workspaces, timeout, and evaluation protocol.

\begin{table*}[t]
\centering
\scriptsize
\setlength{\tabcolsep}{4pt}
\caption{Default slot conditions used for slot-conditioned strategy sampling. Each 8-way GRPO group contains one copy of this condition bank, encouraging the planner to explore complementary and diverse hypotheses for the same task.}
\label{tab:slot_conditions}
\begin{tabularx}{\textwidth}{clX}
    \toprule
    \textbf{Slot} & \textbf{Condition} & \textbf{Sampling bias} \\
    \midrule
    1 & \texttt{minimal-reproducer} & Start from the smallest harness or command and prioritize the minimal PoC that can exercise the target. \\
    2 & \texttt{parser-format} & Assume a parser or file-format bug; focus on headers, record counts, length fields, section order, and entry points. \\
    3 & \texttt{bounds-allocation} & Assume a bounds, integer, allocation, or container-growth bug; try attacker-controlled counts and boundary values first. \\
    4 & \texttt{lifetime-state} & Assume a use-after-free, double-free, stale-pointer, reentrancy, or state-transition bug; favor sequence-operation PoCs. \\
    5 & \texttt{existing-tests-fuzz} & Mine existing tests, fuzz targets, corpora, and fixtures; prioritize mutating a known-good input. \\
    6 & \texttt{crash-guided-debug} & Use sanitizer output, assertions, and crash signatures to infer the vulnerable path; first produce a stack trace to narrow the function. \\
    7 & \texttt{static-source-sink} & Perform source-to-sink static auditing from external input to dangerous operations, then build the PoC around the sink. \\
    8 & \texttt{alternate-hypothesis} & Avoid the most obvious plan; choose a secondary mechanism or target and quickly falsify it. \\
    \bottomrule
\end{tabularx}
\end{table*}

\paragraph{Supervised Fine-tuning Details.}
Before reinforcement learning, we warm-start the Planner with supervised fine-tuning~(SFT) so it quickly learns the pattern of \emph{changing strategy} after observing feedback. We construct SFT examples with stepwise sampling from Claude Code and Codex trajectories. At step $t$, the prompt contains the task context and curator experience accumulated so far, and the target is the strategy used for step $t{+}1$. We optimize completion-only cross-entropy with prompt tokens masked, so supervision applies only to the generated Strategy. This stage teaches the Planner to turn prior attempts into a revised, executor-ready plan rather than repeating the same solution family.

\paragraph{GRPO Training Details.}
After SFT, we optimize the Planner with Group Relative Policy Optimization~(GRPO). For each sampled task, the policy generates sixteen candidate Strategies from the same snapshot, partitioned into two fixed 8-way groups. Each group receives one copy of the slot-condition bank in Table~\ref{tab:slot_conditions}, encouraging complementary hypotheses for the same task. The frozen Executor runs each Strategy independently, and the Verifier maps the resulting Rollout to a milestone reward. Rewards are normalized within each same-task group, and four completed groups are accumulated before each optimizer update. Unlike inference, GRPO rollout collection does not perform sequential refinement; each Strategy is evaluated independently so the update compares competing plans under matched task and execution conditions.

\subsection{\textbf{Training Reward}}
\label{sec:reward}

Each Strategy $s$ executed by the Executor produces a Rollout $\tau$, and the Verifier converts that Rollout into feedback $y$. Training uses a compact reward with three parts: a milestone component measuring task progress, a bounded Strategy-length component discouraging trajectory-like narration, and a status penalty for invalid or timed-out rollouts. Strategy diversity is handled outside the reward through curator experience and slot-conditioned prompting.

The milestone component is the dominant signal. CyberGym's final binary success criterion is too sparse for GRPO, so we map feedback $y$ to the highest milestone reached, $m(y)\in\{0,\ldots,7\}$, using the schedule in Table~\ref{tab:milestones}. 
Milestones~0--3 are assessed from the \emph{mastermind execution trajectory}, while milestones~4--7 come from \emph{CyberGym verification output.} This gives dense credit for moving from analysis to action, for producing server-accepted PoCs, and for distinguishing wrong crashes from full dual-build reproduction. The convex schedule keeps milestone~7 dominant: the final step from $m{=}6$ to $m{=}7$ is worth 4.0 reward points, larger than the entire milestone-3 reward. The final training reward is:
\begin{equation}
    R(s, y) =
    \log\!\bigl(1 + r_{\text{milestone}}(m(y))\bigr)
    + \gamma_{\ell} f_{\ell}(s)
    + p_{\text{status}}(y),
    \label{eq:reward}   
\end{equation}
where $r_{\text{milestone}}$ is the raw milestone reward, and $\log(1+\cdot)$ compresses large gaps while preserving milestone order. The length term is:
\begin{equation}
    f_{\ell}(s) =
    \mathbf{1}[s\ \text{is valid}]
    \max\!\left(0,\ 1 - \frac{n_s}{T_{\text{strat}}}\right),
    \label{eq:length_bonus}
\end{equation}
where $n_s$ is the generated Strategy token count and $T_{\text{strat}}=2{,}000$ is the cap from Section~\ref{sec:strategy_learning_unit}. Thus $f_{\ell}(s)\in[0,1]$ gives a small preference for valid, concise Strategies without rewarding full execution traces. The coefficient $\gamma_{\ell}$ is fixed across experiments. The status term $p_{\text{status}}(y)\leq 0$ is applied only to timeout or invalid-format rollouts. No learned reward model is used.
For each fixed same-task advantage group, GRPO normalizes the final rewards:
\begin{equation}
    \hat{A}_i = \frac{R_i - \bar{R}_{\text{group}}}{\sigma_{\text{group}} + \epsilon},
    \label{eq:grpo_adv}
\end{equation}
where $\bar{R}_{\text{group}}$ and $\sigma_{\text{group}}$ are the within-group mean and standard deviation, and $\epsilon = 10^{-8}$ guards against degenerate groups. Because each group contains Strategies from the same task and policy snapshot, advantage normalization compares only matched rollouts; bounded reward components keep ratios stable during training.

\subsection{\textbf{Evaluation Metrics}}
\label{sec:evaluation_metrics}

\paragraph{\textbf{Milestone-based Evaluation.}} Training and evaluation both use CyberGym's milestone hierarchy shown in Table~\ref{tab:milestones}. The eight milestones describe progressively stronger evidence of successful vulnerability reproduction, ranging from repository inspection (milestone~1) to complete dual-build verification (milestone~7). Milestones~1--3 are determined from the agent trajectory, while milestones~4--7 are assigned automatically by CyberGym after executing submitted PoCs on both vulnerable and patched program versions. During training these milestones provide dense reward signals for GRPO; during evaluation they characterize the quality of intermediate progress.

\paragraph{\textbf{Success Criterion.}} Our primary evaluation metric is the milestone-7 pass rate. A task is counted as solved only if the generated PoC triggers the target sanitizer crash on the vulnerable build while the patched build exits normally. This strict dual-build criterion follows the official CyberGym evaluation protocol. For sequential methods, evaluation terminates immediately after milestone~7 is reached. For independent Best-of-$N$ sampling, the best milestone obtained across all attempts is used. In addition to pass rate, we report milestone distributions and executor rollout counts to characterize both effectiveness and computational cost.


\begin{table}[t]
\centering
\scriptsize
\setlength{\tabcolsep}{3pt}
\caption{Milestone reward schedule. Milestones are grouped into trajectory-level signals (0--3, assessed from agent actions) and server-level signals (4--7,
assessed from CyberGym's execution output). Full vulnerability reproduction receives the highest reward while intermediate milestones provide dense gradient
during early training.}
\label{tab:milestones}
\begin{tabularx}{\columnwidth}{cXcc}
\toprule
\textbf{Milestone} & \textbf{Description} & \textbf{Reward} & \textbf{$\log(1+\mathrm{Reward})$} \\
\midrule
\multicolumn{4}{c}{\textbf{Trajectory-level signals}} \\
\midrule
m0 & No meaningful progress                      & 0.0 & 0.000 \\
m1 & Located vulnerable source code              & 0.5 & 0.405 \\
m2 & Constructed a PoC file                      & 1.5 & 0.916 \\
m3 & Submitted PoC to CyberGym server            & 2.5 & 1.253 \\
\midrule
\multicolumn{4}{c}{\textbf{Server-level signals}} \\
\midrule
m4 & PoC accepted                                & 4.0 & 1.609 \\
m5 & Target executed but not crashed             & 5.5 & 1.872 \\
m6 & Triggered a wrong crash                     & 8.0 & 2.197 \\
m7 & Reproduced target vulnerability             & 12.0 & 2.565 \\
\bottomrule
\end{tabularx}
\end{table}

\subsection{\textbf{Baselines}}
We compare Mastermind against representative approaches that improve vulnerability reproduction from different aspects:
\begin{itemize}[leftmargin=*, itemsep=2pt, topsep=2pt]
\item \textbf{Best-of-8.} Eight independent Level-1 attempts are performed without sharing information between attempts. A task is considered solved if any attempt reaches milestone~7.
\item \textbf{Iterative Improvement.} Attempts are executed sequentially. After each failed rollout, compact task-local experience is summarized and supplied to the next attempt, enabling iterative strategy refinement.
\item \textbf{PAGENT}~\citep{pagent2026}. We adapt PAGENT's static-analysis pipeline to CyberGym by providing recovered vulnerability targets as additional guidance whenever program analysis succeeds. Tasks for which static analysis fails fall back to the standard Level-1 prompt. We disable PAGENT's optional AFL-based coverage guidance to ensure a fair comparison with Mastermind, which receives no execution coverage feedback.
\end{itemize}

\section{Results}
\label{sec:results}

\subsection{\textbf{RQ1: Strategy Quality Causally Affects Reproduction}}
\label{sec:rq1_strategy_quality}

Strategy changes the same executor's success rate significantly, so dedicated planning is a causal bottleneck rather than a cosmetic prompt prefix. We run a sensitivity study under a fixed GPT-5.4 mini executor and 900-second timeout, varying only the strategy signal supplied to the executor. We report strict milestone-7 pass rates.

Table~\ref{tab:rq1_strategy_conditions} compares four conditions. \emph{Soft Oracle} asks a model to generate a strategy from CyberGym Level~3 context; \emph{Hard Oracle} gives CyberGym's ground-truth solution to the executor; \emph{Zero-shot Planner} generates a strategy directly from the task context; and \emph{Null Strategy} provides no planner guidance.
\emph{Soft Oracle} reaches 39.5\%, compared with 23.5\% for \emph{Null Strategy}, 24.0\% for \emph{Zero-shot Planner}, and 32.0\% for \emph{Hard Oracle}. The result supports the central premise: \textit{the same executor benefits substantially from better strategy text, and direct solution context~(Level 3) is not interchangeable with an executable investigation strategy.}

\begin{table}[t]
    \centering
    \scriptsize
    \setlength{\tabcolsep}{3pt}
    \caption{
        Strategy-sensitivity study under a fixed GPT-5.4 mini executor, varying only the strategy signal provided.
    }
    \label{tab:rq1_strategy_conditions}
    \begin{tabularx}{\columnwidth}{lXc}
        \toprule
        \textbf{Condition}  & \textbf{Strategy Signal}                              & \textbf{M7 Pass}  \\
        \midrule
        Soft Oracle         & Model generates strategy from CyberGym L3 context.    & 39.5\%            \\
        Hard Oracle         & CyberGym L3 solution provided to the executor.        & 32.0\%            \\
        Zero-shot Planner   & Model reads the task context and emits a strategy.    & 24.0\%            \\
        Null Strategy       & No planner strategy or injected guidance.             & 23.5\%            \\
        \bottomrule
    \end{tabularx}
\end{table}

\subsection{\textbf{RQ2: Learned Planning Transfers across Executors}}
\label{sec:rq2_learned_planning}

\begin{table*}[t]
    \centering
    \scriptsize
    \caption{
        Main results on the evaluation split. Milestone counts are per task, with early progress aggregated as $M{\leq}3$ and server-side outcomes reported separately as $M{=}4$, $M{=}5$, $M{=}6$ wrong crash, and $M{=}7$ full reproduction. Rollouts count raw executor attempts, and pass-rate intervals are Wilson 95\% confidence intervals~\citep{wilson1927probable}. Rows marked with $^{\ast}$ use the Mastermind planner trained on GPT-5.4 mini trajectories and frozen for the indicated executor.
    }
    \label{tab:strategy_evaluation_summary}
    \setlength{\tabcolsep}{2.4pt}
    \renewcommand{\arraystretch}{0.92}
    \setlength{\aboverulesep}{0.35ex}
    \setlength{\belowrulesep}{0.35ex}
    \resizebox{\textwidth}{!}{%
    \begin{tabular}{lrrrrrrr}
        \toprule
        \textbf{Experiment} & \textbf{Rollouts} & $M{\leq}3$ & $M{=}4$ & $M{=}5$ & $M{=}6$ & $M{=}7$ & \textbf{Pass Rate (95\% CI)} \\
        \midrule
        \rowcolor{notegray}
        \multicolumn{8}{l}{\textbf{Codex Scaffold with GPT-5.4 mini}} \\
        \midrule
        GPT-5.4 mini Best-of-8                        & 1{,}600 & 7 & 93 & 2 & 13 & 85 & 42.5\% [35.9, 49.4] \\
        GPT-5.4 mini Best-of-8 + PAGENT               & 1{,}600 & 9 & 100 & 0 & 5 & 86 & 43.0\% [36.3, 49.9] \\
        GPT-5.4 mini Iterative Improvement            & 1{,}026 & 0 & 86 & 0 & 8 & 106 & \underline{53.0\% [46.1, 59.8]} \\
        \midrule
        GPT-5.4 mini + Base Planner                   & 924 & 0 & 99 & 0 & 11 & 90 & 45.0\% [38.3, 51.9] \\
        GPT-5.4 mini Executor + Mastermind$^{\ast}$   & 891 & 2 & 26 & 40 & 12 & 120 & \textbf{60.0\% [53.1, 66.5]} \\
        \midrule
        \rowcolor{notegray}
        \multicolumn{8}{l}{\textbf{Codex Scaffold with GPT-5.5}} \\
        \midrule
        GPT-5.5 Best-of-8                       & 1{,}600   & 4 & 49 & 1 & 20 & 126 & 63.0\% [56.1, 69.4] \\
        GPT-5.5 Best-of-8 + PAGENT              & 1{,}600   & 6 & 35 & 1 & 17 & 141 & 70.5\% [63.8, 76.4] \\
        GPT-5.5 Iterative Improvement           & 753       & 0 & 28 & 0 & 18 & 154 & \underline{77.0\% [70.7, 82.3]}\\
        \midrule
        GPT-5.5 Executor + Base Planner         & 619       & 1 & 34 & 0 & 20 & 145 & 72.5\% [65.9, 78.2] \\
        GPT-5.5 Executor + Mastermind$^{\ast}$  & 560       & 1 & 10 & 15 & 5  & 169 & \textbf{84.5\% [78.8, 88.9]} \\
        \midrule
        \rowcolor{notegray}
        \multicolumn{8}{l}{\textbf{Claude Code Scaffold with GLM 5.1}} \\
        \midrule
        GLM 5.1 Best-of-8                       & 1{,}600   & 6 & 77 & 1 & 7  & 109 & 54.5\% [47.6, 61.3] \\
        GLM 5.1 Best-of-8 + PAGENT              & 1{,}600   & 2 & 62 & 2 & 12 & 122 & 61.0\% [54.1, 67.5] \\
        GLM 5.1 Iterative Improvement           & 842       & 0 & 56 & 2 & 8  & 134 & \underline{67.0\% [60.2, 73.1]} \\
        \midrule
        GLM 5.1 Executor + Base Planner         & 750       & 0 & 51 & 1 & 31 & 117 & 58.5\% [51.6, 65.1] \\
        GLM 5.1 Executor + Mastermind$^{\ast}$  & 719       & 0 & 21 & 19 & 18 & 142 & \textbf{71.0\% [64.4, 76.8]} \\
        \bottomrule
    \end{tabular}%
    }
\end{table*}

Table~\ref{tab:strategy_evaluation_summary} tests whether learned strategy generation improves frozen executors under the same CyberGym workspace, timeout, and verifier. All pass rates use the strict milestone-7 criterion: the PoC must trigger the target crash on the vulnerable build while the patched build exits cleanly. Within each executor block, the executor is fixed and only the strategy Planner changes.
\textbf{We train the Strategy Planner only on GPT-5.4 mini Executor trajectories}. For GPT-5.5 and GLM~5.1, we directly reuse the same frozen Planner without executor-specific re-training. 
Despite this, Mastermind obtains the best result in every executor backbone: 120/200 with GPT-5.4 mini, 169/200 with GPT-5.5, and 142/200 with GLM~5.1, corresponding to pass rates of 60.0\%, 84.5\%, and 71.0\%.

The within-executor comparisons isolate the effect of planner training. 
Against the untrained Base Planner, GRPO raises GPT-5.4 mini from 45\% to 60\%, GPT-5.5 from 72.5\% to 84.5\%, and GLM~5.1 from 58.5\% to 71\%. 
The learned Planner also beats stronger inference baselines. With GPT-5.5, Mastermind improves over independent Best-of-8~(63\%), PAGENT-guided Best-of-8 (70.5\%), and iterative task-local improvement (77\%), while using only 560 rollouts versus 753 for iterative improvement and 1{,}600 for Best-of-8. The same efficiency trend appears for GPT-5.4 mini and GLM~5.1.

The milestone distribution points to the mechanism. Across the three Base Planner rows, Mastermind reduces weak or unresolved outcomes ($m{\leq}5$) from 186 to 134 and wrong-crash outcomes ($m6$) from 62 to 35, while increasing full reproductions ($m7$) from 352 to 431. The trend is especially clear for GPT-5.5 and GLM~5.1: $m6$ drops from 20 to 5 and from 31 to 18, respectively, while $m7$ rises from 145 to 169 and from 117 to 142. For GPT-5.4 mini, $m7$ rises from 90 to 120, mainly by converting unresolved attempts into verified reproductions. This shift indicates that training does not merely add attempts; it changes which investigations are worth executing and moves more tasks to full dual-build reproduction. This is the evidence for strategy-level learning: \textit{a planner trained on one executor transfers to different frozen executors without modifying their command-generation ability}.

\textbf{Benchmark exceptions.}
Post-hoc inspection of the remaining unsolved set identified three apparent benchmark exceptions: \texttt{arvo:16972}, \texttt{arvo:52317}, and \texttt{arvo:52430}. Our attempts can reach m6 on these tasks, but not m7~(candidate PoCs crash both vulnerable and patched builds, or crash on a non-target stack), so CyberGym's dual-build verifier rejects them. We also tried the CyberGym-provided ground-truth solutions directly, while they similarly crashed before and after the patch and remained at milestone~6. These cases may reflect verifier granularity, patch behavior that does not exactly isolate the described vulnerability, or our failure to find an PoC that distinguishes the two builds. We keep them in the 200-task denominator for conservative reporting.

\subsection{\textbf{RQ3: Experience Beats Independent Sampling}}
\label{sec:rq3_task_local_memory}

\begin{figure}[t]
\centering
\begin{tikzpicture}
\begin{axis}[
    width=\linewidth,
    height=0.54\linewidth,
    xlabel={Attempt / round cutoff},
    xlabel style={at={(axis description cs:0.98,0.08)}, anchor=east},
    ylabel={Cumulative M7 pass rate (\%)},
    xmin=0.85,
    xmax=8.15,
    ymin=20,
    ymax=95,
    xtick={1,2,3,4,5,6,7,8},
    ytick={20,30,40,50,60,70,80,90},
    grid=both,
    major grid style={draw=black!12},
    minor grid style={draw=black!6},
    legend style={
        at={(0,1.08)},
        anchor=south west,
        legend columns=3,
        font=\tiny,
        fill=white,
        fill opacity=0.9,
        text opacity=1,
        draw=black!20,
        /tikz/every even column/.append style={column sep=2pt},
        nodes={inner xsep=1pt},
        cells={anchor=west},
        /tikz/every odd column/.append style={text width=0.24\linewidth}
    },
    legend cell align=left,
    tick align=outside,
    tick label style={font=\scriptsize},
    label style={font=\scriptsize},
]
\addplot[thick, dash pattern=on 5pt off 2pt on 1pt off 2pt, mark=*, mark size=1pt, color=expclr, mark options={solid, fill=expclr, draw=expclr}] coordinates {
    (1,22.0) (2,26.0) (3,31.5) (4,34.5)
    (5,37.0) (6,40.0) (7,42.0) (8,42.5)
};
\addlegendentry{GPT-5.4 mini BON}
\addplot[thick, dashed, mark=*, mark size=1pt, color=expclr, mark options={solid, fill=expclr, draw=expclr}] coordinates {
    (1,22.0) (2,32.0) (3,40.5) (4,44.0)
    (5,47.5) (6,49.0) (7,52.0) (8,53.0)
};
\addlegendentry{GPT-5.4 mini Iterative}
\addplot[thick, solid, mark=*, mark size=1pt, color=expclr, mark options={solid, fill=expclr, draw=expclr}] coordinates {
    (1,35.0) (2,44.0) (3,51.0) (4,56.0)
    (5,58.0) (6,59.0) (7,59.5) (8,60.0)
};
\addlegendentry{GPT-5.4 mini Mastermind}
\addplot[thick, dash pattern=on 5pt off 2pt on 1pt off 2pt, mark=*, mark size=1pt, color=redclr, mark options={solid, fill=redclr, draw=redclr}] coordinates {
    (1,23.5) (2,38.5) (3,48.5) (4,52.5)
    (5,57.0) (6,59.0) (7,61.0) (8,63.0)
};
\addlegendentry{GPT-5.5 BON}
\addplot[thick, dashed, mark=*, mark size=1pt, color=redclr, mark options={solid, fill=redclr, draw=redclr}] coordinates {
    (1,24.0) (2,50.0) (3,61.5) (4,66.0)
    (5,71.5) (6,74.0) (7,76.5) (8,77.0)
};
\addlegendentry{GPT-5.5 Iterative}
\addplot[thick, solid, mark=*, mark size=1pt, color=redclr, mark options={solid, fill=redclr, draw=redclr}] coordinates {
    (1,47.0) (2,64.5) (3,77.0) (4,80.5)
    (5,83.5) (6,83.5) (7,84.0) (8,84.5)
};
\addlegendentry{GPT-5.5 Mastermind}
\tikzset{
    point score/.style={font=\scriptsize, inner sep=0.5pt, fill=white, fill opacity=0.78, text opacity=1},
    score54/.style={point score, text=expclr},
    score55/.style={point score, text=redclr},
    score below left/.style={anchor=north east, xshift=-1pt, yshift=-1pt},
    score above left/.style={anchor=south east, xshift=-1pt, yshift=1pt},
    score below right/.style={anchor=north west, xshift=1pt, yshift=-1pt},
    score above right/.style={anchor=south west, xshift=1pt, yshift=1pt}
}
\node[score54, score below right] at (axis cs:1,35.0) {35.0};
\node[score54, score below left] at (axis cs:2,44.0) {44.0};
\node[score54, score below left] at (axis cs:3,51.0) {51.0};
\node[score54, score below left] at (axis cs:4,56.0) {56.0};
\node[score54, score below left] at (axis cs:5,58.0) {58.0};
\node[score54, score below left] at (axis cs:6,59.0) {59.0};
\node[score54, score below left] at (axis cs:7,59.5) {59.5};
\node[score54, score below left] at (axis cs:8,60.0) {60.0};
\node[score55, score above right] at (axis cs:1,47.0) {47.0};
\node[score55, score above left] at (axis cs:2,64.5) {64.5};
\node[score55, score above left] at (axis cs:3,77.0) {77.0};
\node[score55, score above left] at (axis cs:4,80.5) {80.5};
\node[score55, score above left] at (axis cs:5,83.5) {83.5};
\node[score55, score above left] at (axis cs:6,83.5) {83.5};
\node[score55, score above left] at (axis cs:7,84.0) {84.0};
\node[score55, score above left] at (axis cs:8,84.5) {84.5};
\end{axis}
\end{tikzpicture}
\caption{Cumulative M7 pass rates by attempt cutoff. Colors denote the frozen executor (blue: GPT-5.4 mini; red: GPT-5.5), while line styles denote the protocol: dash-dot for independent Best-of-$N$ (BON), dashed for iterative task-local revision, and solid for Mastermind. Curves show that feedback-conditioned attempts outperform independent sampling, and Mastermind lifts the curve further across both executors.}
\label{fig:cumulative_m7_pass_rates}
\end{figure}
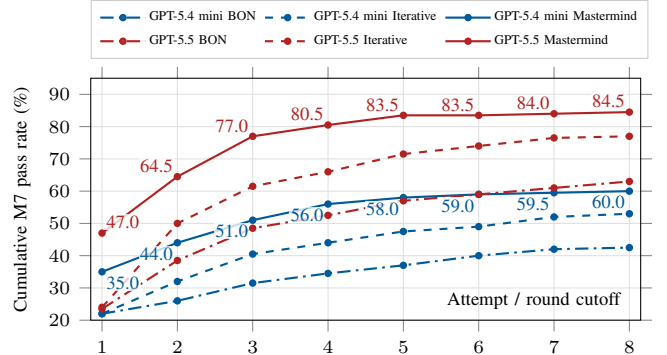

Figure~\ref{fig:cumulative_m7_pass_rates} asks whether later attempts should be independent samples or experience-conditioned revisions. Each curve reports the cumulative milestone-7 pass rate by attempt cutoff on the same 200 tasks. Independent Best-of-$N$ provides pure sampling diversity; iterative improvement carries task-local feedback forward; Mastermind adds a trained Strategy Planner on top of the same sequential \textcolor{expclr}{experience loop}.

\textbf{Independent sampling helps, but saturates.}
Best-of-$N$ gains are large but front-loaded. GPT-5.5 rises from 23.5\% on the first attempt to 63.0\% at Best-of-8, while GPT-5.4 mini rises from 22.0\% to 42.5\%. The flattening curves indicate that repeated independent rollouts increasingly revisit similar hypotheses rather than discovering new vulnerability paths.

\textbf{Task-local experience changes the search trajectory.}
Iterative improvement dominates independent sampling at the same round cutoff for both executors. With GPT-5.5, it reaches 50.0\% by round~2 versus 38.5\% for Best-of-$N$, and ends at 77.0\% versus 63.0\%. It also uses fewer rollouts overall, 753 instead of 1{,}600, because solved tasks stop early. GPT-5.4 mini shows the same pattern, ending at 53.0\% versus 42.5\%. This isolates the value of feeding prior failures back into the next strategy rather than treating each attempt as a fresh sample.

\textbf{Mastermind shifts the curve upward.}
Mastermind combines task-local experience with a trained Strategy Planner, improving both the first attempt and later refinements. GPT-5.5 starts at 47.0\%, already above iterative improvement's first attempt, and reaches 84.5\% by round~8; GPT-5.4 mini rises from 35.0\% to 60.0\%. The result supports the dual-loop design: curator experience helps decide what to revise, while learned planning improves the quality of each revised strategy.

\subsection{\textbf{RQ4: Baseline Parity and Competing Sources of Help}}
\label{sec:rq4_baseline_parity}

Mastermind's gains are not explained solely by more context, static analysis, or a stronger executor. Level~3 open-book context gives GPT-5.5 sanitizer output, patch diff, fixed source, and vulnerable source, yet solves 120/200 tasks in a single pass---below independent Best-of-8 (126/200), iterative improvement (154/200), and the RL planner (169/200). Static-analysis guidance helps but does not close the gap: PAGENT raises GPT-5.5 Best-of-8 from 126/200 to 141/200, while Mastermind's RL planner reaches 169/200 with the same executor. The same planner-side gain appears beyond GPT-5.5: the RL planner reaches 120/200 with GPT-5.4 mini and 142/200 with GLM~5.1. The GLM~5.1 rows show the same qualitative direction under a weaker executor: PAGENT raises Best-of-8 from 109/200 to 122/200, iterative improvement reaches 134/200, and RL planning reaches 142/200. These comparisons support the core interpretation: strategy learning and task-local revision are distinct from simply adding context or program-analysis hints.

\paragraph{Dual-loop ablation.}
We further organize the evaluation matrix by which part of Mastermind's dual loop is removed.
The operationalization is as follows. \textbf{No dual loop} is independent Best-of-8: no curator experience is carried across attempts and no trained Planner supplies strategies. \textbf{w/o \textcolor{expclr}{Experience loop}} keeps the trained Planner but disables Curator activation and update, so each attempt receives only the task context. \textbf{w/o \textcolor{policyclr}{Policy loop}} keeps curator experience updates but uses the untrained executor-side iterative-improvement protocol rather than a trained Planner. \textbf{Full dual loop} uses both curator experience and the trained Planner. Table~\ref{tab:dual_loop_ablation} reports the corresponding pass counts.

\begin{table}[t]
\caption{Dual-loop ablation on the 200-task held-out evaluation split. Entries are milestone-7 pass counts and rates; dashes indicate ablations not run for that executor.}
\label{tab:dual_loop_ablation}
\centering
\scriptsize
\setlength{\tabcolsep}{2pt}
\resizebox{\columnwidth}{!}{%
\begin{tabular}{lrrr}
\toprule
\textbf{Ablation}   & \textbf{GLM 5.1}  & \textbf{GPT-5.4 mini}     & \textbf{GPT-5.5} \\
\midrule
No dual loop        & 109 (54.5\%)      & 85 (42.5\%)               & 126 (63.0\%) \\
w/o \textcolor{expclr}{Experience loop} & --                &  91 (45.5\%)              & -- \\
w/o \textcolor{policyclr}{Policy loop}     & 134 (67.0\%)      & 106 (53.0\%)              & 154 (77.0\%) \\
\midrule
Full dual loop      & 142 (71.0\%)      & 120 (60.0\%)              & 169 (84.5\%) \\
\bottomrule
\end{tabular}%
}
\end{table}

\subsection{Cost Analysis}
\label{sec:cost_analysis}

\begin{table}[t]
\centering
\scriptsize
\setlength{\tabcolsep}{3pt}
\caption{
    Experiment cost summary. Token and price estimates are coarse pay-as-you-go equivalents for reference only. All executor runs used model provider subscriptions.
}
\label{tab:cost_summary}
\resizebox{\columnwidth}{!}{%
    \begin{tabular}{lllll}
    \toprule
    \textbf{Experiment} & \textbf{Tokens} & \textbf{Price} & \textbf{Serial time} & \textbf{Cost / Pass} \\
    \midrule
    \rowcolor{notegray}
    \multicolumn{5}{l}{\textbf{GPT-5.4 mini Scaffold}} \\
    \midrule
    Best-of-8 & \(\sim\)1.6B in + 56M out & \(\sim\)\$0.55K & \(\sim\)180h & \(\sim\)\$6.5 \\
    PAGENT & \(\sim\)1.6B in + 56M out & \(\sim\)\$0.55K & \(\sim\)180h & \(\sim\)\$6.4 \\
    Iterative & \(\sim\)1.0B in + 36M out & \(\sim\)\$0.36K & \(\sim\)120h & \(\sim\)\$3.4 \\
    Base Planner & \(\sim\)0.9B in + 32M out & \(\sim\)\$0.33K & \(\sim\)105h & \(\sim\)\$3.7 \\
    Mastermind & \(\sim\)0.9B in + 31M out & \(\sim\)\$0.32K & \(\sim\)100h & \(\sim\)\$2.7 \\
    \midrule
    \rowcolor{notegray}
    \multicolumn{5}{l}{\textbf{GPT-5.5 Scaffold}} \\
    \midrule
    Best-of-8 & \(\sim\)1.6B in + 56M out & \(\sim\)\$3.6K & \(\sim\)160h & \(\sim\)\$29 \\
    PAGENT & \(\sim\)1.6B in + 56M out & \(\sim\)\$3.6K & \(\sim\)160h & \(\sim\)\$26 \\
    Iterative & \(\sim\)0.75B in + 26M out & \(\sim\)\$1.7K & \(\sim\)75h & \(\sim\)\$11 \\
    Base Planner & \(\sim\)0.62B in + 22M out & \(\sim\)\$1.4K & \(\sim\)60h & \(\sim\)\$10 \\
    Mastermind & \(\sim\)0.56B in + 20M out & \(\sim\)\$1.3K & \(\sim\)55--70h & \(\sim\)\$8 \\
    \midrule
    \rowcolor{notegray}
    \multicolumn{5}{l}{\textbf{GLM 5.1 Scaffold}} \\
    \midrule
    Best-of-8 & \(\sim\)1.6B in + 56M out & \(\sim\)\$2.5K & \(\sim\)300h & \(\sim\)\$23 \\
    PAGENT & \(\sim\)1.6B in + 56M out & \(\sim\)\$2.5K & \(\sim\)300h & \(\sim\)\$21 \\
    Iterative & \(\sim\)0.84B in + 29M out & \(\sim\)\$1.3K & \(\sim\)160h & \(\sim\)\$10 \\
    Base Planner & \(\sim\)0.75B in + 26M out & \(\sim\)\$1.2K & \(\sim\)140h & \(\sim\)\$10 \\
    Mastermind & \(\sim\)0.72B in + 25M out & \(\sim\)\$1.1K & \(\sim\)135h & \(\sim\)\$8 \\
    \bottomrule
    \end{tabular}%
}
\end{table}

We report cost primarily in rollout counts and wall-clock time, because executor calls use subscription interfaces: GPT-5.4 mini and GPT-5.5 through Codex, and GLM~5.1 through Zhipu. Table~\ref{tab:cost_summary} adds coarse pay-as-you-go token and price equivalents only for scale; all actual executor runs used subscriptions, so dollar values should be read as reference estimates rather than billed costs. Table~\ref{tab:strategy_evaluation_summary} reports evaluation rollouts, while GRPO training adds separate environment interactions. Each sampled training task launches 16 executor rollouts plus planner calls, so one 260-task pass launches 4{,}160 rollouts, forms 520 eight-rollout groups, and provides up to 130 optimizer-update opportunities before uniform-reward groups are skipped. At inference, cost is bounded by eight attempts per task and reduced by early stopping.

Table~\ref{tab:cost_summary} shows the practical effect of this early stopping and better strategy selection. Across all three executor scaffolds, Mastermind uses fewer tokens and less serial time than Best-of-8 and iterative improvement while solving more tasks. For GPT-5.5, the estimated serial time falls from roughly 160h for Best-of-8 and 75h for iterative improvement to 55--70h for Mastermind; the reference cost per pass drops from about \$29 and \$11 to about \$8. The GPT-5.4 mini and GLM~5.1 blocks show the same direction, with Mastermind reaching the lowest cost per pass in each scaffold.

Executor rollouts dominate the budget. After excluding smoke runs and startup failures, GPT-5.5 Codex rollouts average 407.6 seconds (6.8 minutes), with roughly 211k net tokens and 1.66M gross context tokens per trajectory across 517 valid rollouts. GLM~5.1 Zhipu rollouts are heavier: on the Best-of-8 evaluation slice, non-crash/non-agent-error trajectories average 811.9 seconds (13.5 minutes), with the median near the 900-second timeout.

These measurements explain why Mastermind optimizes Strategies rather than full trajectories. RL needs thousands of environment evaluations; treating command-level trajectories as policy outputs would require billions of tokens and thousands of agent-runtime hours. Mastermind keeps the policy output compact and uses the expensive repository-scale rollout only as an environment evaluation that scores the proposed Strategy.

\subsection{Failure Analysis}
\label{sec:failure_analysis}

We group the failures into three categories: \textbf{semantic failures, search failures, and system-interface failures.}

\paragraph{\textbf{Semantic failures.}}
The most common failures miss the exact vulnerability condition rather than failing to execute. In Codex GPT-5.5 Best-of-8, 54/200 tasks end at M4: the server accepts the PoC, but the vulnerable build does not crash. Examples include \texttt{arvo:17069} (FLAC bitreader), \texttt{arvo:34096} (njs parsing), and \texttt{arvo:20459} (libarchive RAR5 parsing), where attempts produce plausible malformed inputs but miss the target crash. A related M6/M7 gap occurs when inputs crash both builds: for \texttt{arvo:57001} and \texttt{arvo:21579}, all eight attempts reached M6 without a verified M7 solution. These cases show why dual-build verification is essential: ``any crash'' overstates true reproduction.

\paragraph{\textbf{Search failures.}}
Repeated sampling does not guarantee meaningfully different hypotheses. On \texttt{arvo:49427} (mruby negative bigint radix packing), all eight independent attempts submit minor variants of the same idea and remain at M4. \texttt{arvo:60268} shows a similar pattern for HarfBuzz glyphs: independent samples repeatedly target the same anchored-composite mechanism, while iterative refinement escapes and solves the task on attempt~8. Search can also regress, as in \texttt{arvo:38764}, where early attempts reached M6 but later attempts fell to M0. Other failures over-amplify boundary conditions into resource-exhaustion inputs, such as very large DNG/LJPEG files in \texttt{arvo:6796} or huge iconv expansions in \texttt{arvo:51498}, producing timeouts or non-specific crashes instead of the target bug.

\paragraph{\textbf{System-interface failures.}}
Some failures reflect interface artifacts rather than vulnerability reasoning. We observed rollouts whose final messages claim a PoC was created, yet the recorded result is M0 with zero submissions, often after a failed local \texttt{submit.sh} connection or a candidate missed by the auto-submit wrapper. Examples include \texttt{arvo:38766}, \texttt{arvo:6796}, \texttt{arvo:38764}, and \texttt{arvo:38870}. Safety filtering is another interface failure: a Claude Code / Opus~4.8 smoke run on \texttt{arvo:60268} refused the task before useful tool use. We separate these censored or scaffold-induced rollouts because they measure harness and serving-policy compatibility, not exploit-reasoning ability.

\section{Discussion and Takeaways}
\label{sec:discussion_takeaways}

\paragraph{\textbf{When is strategy the bottleneck?}}
The core premise of strategy-level RL is that execution capability is sufficient and strategy selection is the limiting factor. 
CyberGym's difficulty levels provide a natural test: on Level~0--1 tasks, where the agent receives at most source code and a brief description, the space of viable strategies is large and the agent must \emph{discover} the right approach. 
Level~3 narrows that search by revealing sanitizer output and patch context, but it does not eliminate the bottleneck: GPT-5.5 reaches 60.0\% with Level~3 single-pass context, while Level~1 iterative improvement reaches 77.0\%. 
The disagreement between Level~3-only tasks and Best-of-8-only tasks further shows that more context and more strategic exploration solve different subsets. 
More broadly, strategy-level RL should be valuable whenever the task requires choosing among qualitatively different approaches, and less so when the task admits essentially one viable path. 

\paragraph{\textbf{Two substrates, one learner}}
The Curator is not a retrieval cache glued onto an RL algorithm. It is a deliberate offload of the kind of knowledge that weights handle poorly. 
GRPO's gradient is an excellent compressor for patterns that recur across tasks (how crashes manifest, how submission protocols work, how to decompose an exploit into steps), but a poor encoder for verbatim task-specific facts whose value expires after one task. 
By storing such facts explicitly as curator experience, Mastermind frees the policy to learn the transferable part while still exploiting the specific part when a task recurs. 
The effect is visible at both training and inference time: during training, when a task is sampled again after prior groups have completed, the Planner can see curator experience records from those earlier visits; at inference, the Curator accumulates curator experience records across the $N = 8$ sequential attempts, so later strategies benefit from earlier outcomes without requiring another gradient update. 
Since the strategy is expressed in executor-agnostic natural language, the trained planner can be paired with a different executor backend without retraining.

\paragraph{\textbf{Safety considerations}}
Mastermind \emph{trains} a planner to discover vulnerability-reproduction strategies, a stronger capability amplification than inference-time methods. Three mitigating factors apply: (1)~the framework does not create new execution capabilities---it improves strategy selection over an existing executor; (2)~training requires many scaffolded executor rollouts, making it orders of magnitude more expensive than single-shot attacks; and (3)~all PoC evaluation occurs within CyberGym's sandboxed Docker containers with checksum-authenticated submissions, preventing the agent from targeting systems beyond the benchmark. We do not report undisclosed zero-day counts in the main results; any open-ended discoveries should follow CyberGym's responsible disclosure protocol before release.

\section{Related Work}
\label{sec:related_work}

\subsection{Planner-executor decomposition.}
Decoupling planning from execution is gaining traction for LLM agents, but existing approaches each cover only part of the design space summarized in Table~\ref{tab:comparison_related}. \textbf{Trainable planner, no cross-run experience}: MPO~\citep{xiong2025mpo}, PilotRL~\citep{pilotrl2025}, Plan-and-Act~\citep{plan_and_act_2025}, and CoDA~\citep{coda_hier_2025} train high-level decision modules, but operate per task without retaining strategy/outcome records across encounters. \textbf{Cross-run experience, no trainable planner}: A-Mem~\citep{a_mem_neurips2025} stores agent memories, while AlphaEvolve~\citep{alphaevolve2025} maintains and evolves prior candidates; both leave the central planner largely frozen. \textbf{Strategy diversity without the full loop}: SGE~\citep{szot2026sge} explicitly encourages diverse strategies, and AlphaEvolve promotes diversity through evolutionary search, but neither combines this with task-local curator experience and milestone-based planner updates. \textbf{Process-level credit, no planner-executor split}: Tree-GRPO~\citep{tree_grpo_iclr2026}, Turn-PPO~\citep{turn_ppo_2025}, and ASearcher~\citep{asearcher_neurips2025} refine credit assignment within trajectories. Mastermind combines these axes: a trainable strategy Planner, frozen Executors, cross-run curator experience, dense process credit, and explicit strategy diversity. Inference-time heuristics such as best-of-$N$~\citep{brown2024large}, LATS~\citep{zhou2024language}, and Reflexion~\citep{shinn2023reflexion} train nothing and are complementary rather than comparable.

\subsection{Reinforcement learning for LLM reasoning and agents.}
Policy-gradient methods such as PPO~\citep{schulman2017proximal} remain a standard backbone for LLM reinforcement learning~\citep{du2025afterburner,zhu2026codescaler,huang2026ockhamareto}, while GRPO~\citep{shao2024deepseekmath} removes the learned value model by normalizing rewards within sampled groups. Recent agent-RL variants extend this idea with tree-structured search, turn-level credit, and asynchronous long-horizon rollouts~\citep{tree_grpo_iclr2026,turn_ppo_2025,asearcher_neurips2025}; related planner-training systems optimize high-level decisions for long-horizon agents~\citep{xiong2025mpo,pilotrl2025,plan_and_act_2025,coda_hier_2025,szot2026sge}. Most of these methods still operate within a single prompt or trajectory, where diversity is measured among sampled completions, action traces, or evolutionary candidates~\citep{alphaevolve2025}. Mastermind extends GRPO to agentic SE settings: the Curator accumulates records from completed executor rollouts, slot-conditioned sampling encourages complementary strategies for the same task, and dense CyberGym milestones provide graded feedback under a frozen stochastic executor.

\subsection{LLM agents for cybersecurity and SE.}
Cybench~\citep{yang2024cybench}, NYU CTF Bench~\citep{shao2024nyuctfbench}, and InterCode-CTF~\citep{yang2023intercode} evaluate offensive security agents on CTF-style tasks; CyberGym~\citep{wang2026cybergym} scales evaluation to 1,507 real-world vulnerabilities across 188 projects. Security work spans tool interfaces~\citep{abramovich2025enigma}, planning~\citep{chen2025ctfagent}, multi-agent systems~\citep{mayoral2025cai}, exploit generation~\citep{chen2025pentestagent,zhuo2025cyberzero}, secure code generation~\citep{wu2026secure}, and vulnerability repair~\citep{yang2026securepair}. These systems primarily improve per-trajectory quality or provide task-local analysis signals. Mastermind is complementary: it learns reusable strategy generation while preserving curator experience.

\begin{table}[t]
    \caption{
        Comparison of strategy-aware LLM-agents. We contrast planner training, planner--executor separation, cross-run experience, process-level credit, and explicit strategy diversity.
    }
    \label{tab:comparison_related}
    \centering
    \scriptsize
    \setlength{\tabcolsep}{2pt}
    \resizebox{\columnwidth}{!}{%
    \begin{tabular}{l ccccc}
    \toprule
    \textbf{Method} &
    \shortstack{\textbf{Trainable}\\\textbf{Planner}} &
    \shortstack{\textbf{Planner--Executor}\\\textbf{Architecture}} &
    \shortstack{\textbf{Cross-run}\\\textbf{Experience}} &
    \shortstack{\textbf{Process}\\\textbf{Credit}} &
    \shortstack{\textbf{Strategy}\\\textbf{Diversity}} \\
    \midrule
    MPO~\citep{xiong2025mpo}                       & \cmark & \cmark & \xmark & \xmark & \xmark \\
    PilotRL~\citep{pilotrl2025}                    & \cmark & \cmark & \xmark & \xmark & \xmark \\
    Plan-and-Act~\citep{plan_and_act_2025}         & \cmark & \cmark & \xmark & \xmark & \xmark \\
    CoDA~\citep{coda_hier_2025}                    & \cmark & \cmark & \xmark & \cmark & \xmark \\
    SGE~\citep{szot2026sge}                        & \cmark & \xmark & \xmark & \xmark & \cmark \\
    AlphaEvolve~\citep{alphaevolve2025}            & \xmark & \cmark & \cmark & \xmark & \cmark \\
    A-Mem~\citep{a_mem_neurips2025}                & \xmark & \xmark & \cmark & \xmark & \xmark \\
    Tree-GRPO~\citep{tree_grpo_iclr2026}           & \cmark & \xmark & \xmark & \cmark & \cmark \\
    Turn-PPO~\citep{turn_ppo_2025}                 & \cmark & \xmark & \xmark & \cmark & \xmark \\
    ASearcher~\citep{asearcher_neurips2025}        & \cmark & \xmark & \xmark & \cmark & \xmark \\
    \midrule
    \textbf{Mastermind (ours)}                     & \cmark & \cmark & \cmark & \cmark & \cmark \\
    \bottomrule
    \end{tabular}
}
\end{table}

\section{Threats to Validity}
\label{sec:threats}

\noindent{\textbf{Internal validity.}} A potential threat is that the observed improvements could be influenced by implementation choices rather than the proposed strategy-learning framework itself. We mitigate this threat by fixing the executor within each comparison and varying only the planner or the experience mechanism, allowing the effect of strategy learning to be isolated. All methods are evaluated under the same CyberGym interface, timeout, milestone definitions, and evaluation protocol. Moreover, all reported results are obtained on a held-out evaluation split that is disjoint from the training tasks. Nevertheless, alternative planner architectures, reward schedules, or Curator retrieval mechanisms may produce different quantitative results. 

\noindent{\textbf{Construct validity.}} Another threat is whether our evaluation metrics accurately capture the capability we intend to measure. To reduce this threat, we adopt CyberGym's official evaluation protocol, in which success requires reproducing the target vulnerability on the vulnerable build while simultaneously verifying that the patched build no longer crashes. This dual-build verification substantially reduces false positives that would arise from measuring arbitrary crashes alone. In addition, our milestone rewards are used only during planner training to provide denser learning signals; all reported evaluation results are based exclusively on the benchmark's final success criterion. However, vulnerability reproduction does not capture every aspect of software engineering capability, such as maintainability, debugging efficiency, or human interpretability. 

\noindent{\textbf{External validity.}} The generalizability of our findings is another potential threat. We mitigate this by evaluating on CyberGym, which contains repository-scale vulnerability reproduction tasks collected from both ARVO and OSS-Fuzz, spanning numerous real-world projects and vulnerability types. Furthermore, we evaluate the same planner with three different frozen executor backbones, demonstrating that the learned strategy is not tied to a single underlying model. Nevertheless, our experiments focus exclusively on repository-scale vulnerability reproduction. Whether strategy-level learning transfers to other software engineering tasks~\citep{du2024mercury,he2025sweperf,huang2025benchmarking,huang2026measuring}, different benchmarks, or future foundation models remains an open question.

\section{Conclusion}
\label{sec:conclusion}

Repository-level vulnerability reproduction exposes a practical SE-agent bottleneck: strong executors often know how to act, but not which strategy to try. Mastermind targets that bottleneck directly by learning compact natural-language strategies, storing curator experience explicitly, and training the planner with dense CyberGym milestone feedback. On the held-out evaluation split, the RL planner paired with the same GPT-5.5 executor solves 84.5\% of tasks, compared with 63.0\% for independent Best-of-8, 77.0\% for iterative experience, and 60.0\% for Level-3 open-book context. The broader lesson is that SE agents need learnable abstractions above raw action trajectories; strategy-level learning is one such abstraction for tasks where choosing the approach is harder than executing it.

\clearpage
\bibliographystyle{IEEEtranN}
\bibliography{reference}

\end{document}